\documentclass[10pt,twocolumn,letterpaper]{article}

\usepackage{cvpr}              

\definecolor{cvprblue}{rgb}{0.21,0.49,0.74}
\usepackage[pagebackref,breaklinks,colorlinks,allcolors=cvprblue]{hyperref}

\def\paperID{18} 
\def\confName{CVPR}
\def\confYear{2025}

\title{Dream-Box: Object-wise Outlier Generation for Out-of-Distribution Detection \vspace{-0.5cm}}

\author{Brian K.S. Isaac-Medina\textsuperscript{1},
Toby P. Breckon\textsuperscript{1,2}\\
Department of \{Computer Science\textsuperscript{1}, Engineering\textsuperscript{2}\}, Durham University, Durham, UK
}

\begin{document}
\maketitle

\begin{abstract}
 Deep neural networks have demonstrated great generalization capabilities for tasks whose training and test sets are drawn from the same distribution. Nevertheless, out-of-distribution (OOD) detection remains a challenging task that has received significant attention in recent years. Specifically, OOD detection refers to the detection of instances that do not belong to the training distribution, while still having good performance on the in-distribution task (\eg, classification or object detection). Recent work has focused on generating synthetic outliers and using them to train an outlier detector, generally achieving improved OOD detection than traditional OOD methods. In this regard, outliers can be generated either in feature or pixel space. Feature space driven methods have shown strong performance on both the classification and object detection tasks, at the expense that the visualization of training outliers remains unknown, making further analysis on OOD failure modes challenging. On the other hand, pixel space outlier generation techniques enabled by  diffusion models have been used for image classification using, providing improved OOD detection performance and outlier visualization, although their adaption to the object detection task is as yet unexplored. We therefore introduce Dream-Box, a method that provides a link to object-wise outlier generation in the pixel space for OOD detection. Specifically, we use diffusion models to generate object-wise outliers that are used to train an object detector for an in-distribution task and OOD detection. Our method achieves comparable performance to previous traditional methods while being the first technique to provide concrete visualization of generated OOD objects.
\end{abstract}    
\begin{figure*}[!t]
\centering
\includegraphics[width=\linewidth]{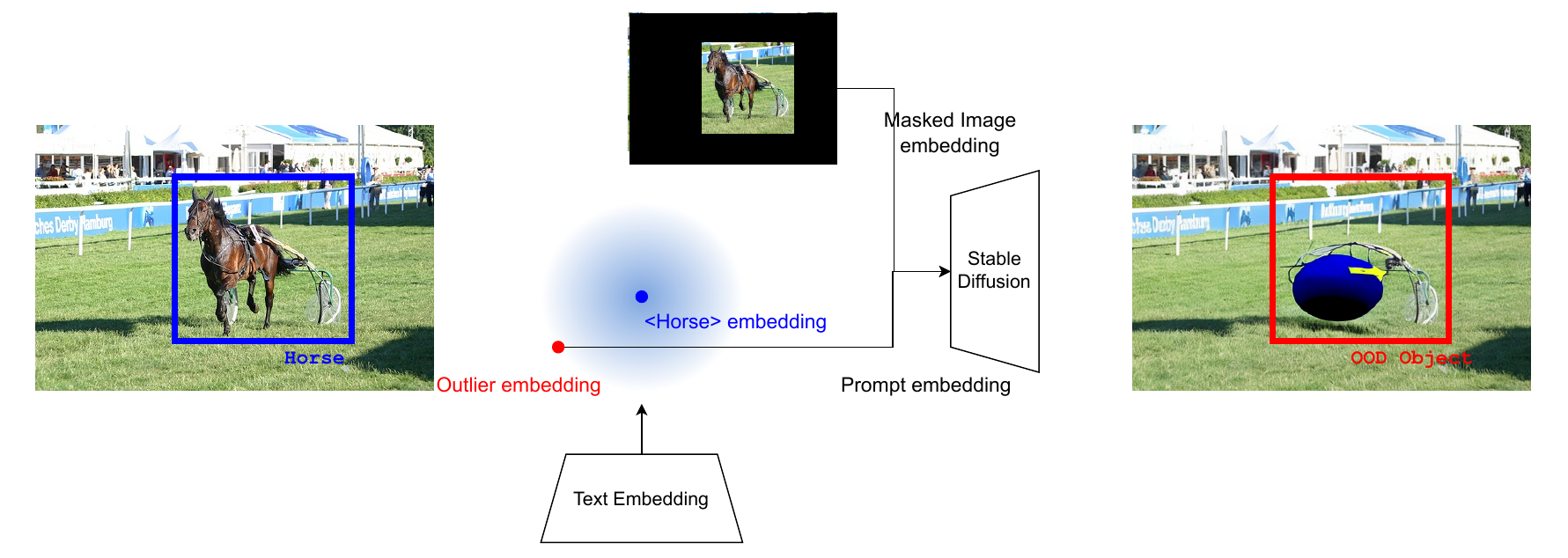}
\caption{Dream-Box enables object-wise OOD detection by generating objects using embeddings far from the class-name text embeddings.}
\label{fig:banner}
\end{figure*}

\section{Introduction}
\label{sec:intro}

Out-of-distribution (OOD) detection has emerged as a critical challenge in the deployment of deep neural networks, particularly in tasks such as classification \cite{he2016deep,liu2021swin,dosovitskiy2020image,pmlr-v162-wortsman22a} and object detection \cite{ren2016faster,wang2023yolov7,zhu2020deformable} in addition to having broader impact potential for practical outlier detection across a range of application tasks \cite{isaac2024towards,gaus25anomaly,gaus23region,gokstorp21saliency, barker22turbine}. While these models exhibit remarkable generalization capabilities within their training distribution, their ability to identify and handle data that deviates from this distribution remains a significant limitation. OOD detection aims to address this issue by distinguishing between in-distribution and out-of-distribution instances, ensuring robust performance in real-world scenarios where the input data may not conform to the training set \cite{yang2024generalized}.

Traditional approaches for OOD detection include using simple softmax probabilities \cite{hendrycks2016baseline} or the Mahalanobis distance \cite{mahalanobis} to leverage the statistical properties of feature distributions, identifying OOD instances by measuring the distance of a given sample from the inlier distribution in the feature space. Another approach consists of using Gram matrices \cite{sastry2020detecting} to capture the correlations between feature maps, providing a robust representation to discriminate between in-distribution and OOD data. A recent approach that has shown great performance for OOD detection is to use the energy score \cite{liu2020energy}, \ie, the \textit{log-sum-exp} operation over the class logits, framing OOD detection as a density estimation problem by assigning lower energy scores to in-distribution samples and higher scores to outliers. Whilst these methods have shown promising results, they often lack the ability to leverage training outliers (\eg, synthetically generated outlier instances). This limitation has spurred the development of outlier generation approaches, which not only improve detection performance but in some cases offer the added benefit of interpretable outlier visualization \cite{du2022vos,du2023dream}.

In terms of outlier synthesis, other works have focused on creating synthetic outliers for training an in-distribution/OOD binary classifier. Among these approaches, feature space outlier generation methods, such as Virtual Outlier Synthesis (VOS) \cite{du2022vos} and Feature Flow Synthesis (FFS) \cite{kumar2023normalizing}, have demonstrated strong performance in both classification and object detection tasks. These techniques generate synthetic outliers in the feature space, enabling the training of outlier classifiers that enhance OOD detection capabilities. However, a notable limitation of these methods is the lack of interpretability and visualization of the generated outliers, which hinders a deeper understanding of their failure modes and limits further analysis towards improved performance. On the other hand, pixel space outlier generation techniques have shown promise in image classification tasks. For instance, Dream-OOD \cite{du2023dream} leverages diffusion models to generate synthetic outlier images directly in pixel space. This technique provides meaningful visualizations of the generated outliers, offering insights into their characteristics, while improving OOD detection. Despite these advancements, the application of pixel space outlier generation to object-wise OOD detection remains largely unexplored.

This work aims to close this gap by introducing a novel approach leveraging diffusion models for object-wise pixel space outlier generation in the context of object detection. Our method generates object-wise outliers (\cref{fig:banner}), which are then used to train an object detector capable of performing both in-distribution tasks and OOD detection. Our method, dubbed Dream-Box, achieves OOD detection performance comparable to state-of-the-art traditional methods while providing interpretable visualizations of OOD objects. This work builds on the foundations laid by Dream-OOD \cite{du2023dream}, which explores the use of generative models for OOD detection, but further extends its applicability to object detection tasks. Our contributions not only advance the state-of-the-art in OOD detection but also open new avenues for research in outlier visualization and analysis, encouraging for more robust and interpretable deep learning systems. 
\\
Reference software code is available at:
\\
\footnotesize
\url{https://github.com/KostadinovShalon/dream-box}
\normalsize
\section{Related Work}
\label{sec:related_work}
Several works investigate OOD detection using different approaches. For instance, some works rely on estimating the probability density of the training data and flagging low-density regions as OOD. For instance Lee \etal \cite{mahalanobis} proposed using the Mahalanobis distance in the feature space of deep neural networks, achieving strong performance in OOD detection tasks. Nonetheless, Mahalanobis distance may not generalize well for complex distributions, such as for object detection. Other methods, such as ODIN \cite{liang2017enhancing} and Gram matrices \cite{sastry2020detecting}, modify pre-trained models to improve OOD detection without retraining. While computationally efficient, these methods often rely on heuristics and may not generalize across diverse datasets. 

A recent approach that has shown promising results uses the free energy score for OOD detection \cite{liu2020energy}. Energy-based models treat OOD detection as a density estimation problem by assigning lower energy scores to in-distribution samples and higher scores to outliers, leveraging the logits of a pre-trained classifier to detect OOD samples, demonstrating high performance on benchmark datasets. Other works have used the energy method while using synthetically generated outliers. For instance, Virtual Outlier Synthesis (VOS) \cite{du2022vos} generates outliers in the feature space by learning Gaussian distributions over the feature representation of different classes and sampling from low-probablity regions. Similarly, Feature Flow Synthesis (FFS) \cite{kumar2023normalizing} learns a reversible transformation from the feature space to a class-agnostic normalized space, where outliers are generated, demonstrating improvement over VOS. 

Additionally, Isaac-Medina \etal \cite{isaac2024towards} extended the VOS and FFS frameworks introducing OLN-SSOS for class-agnostic open-world OOD detection in object detection. In this context, these methods are the first methods to show a significant performance for the object detection task, but often lack interpretability and visualization capabilities. 

\begin{figure*}[t]
\centering
\includegraphics[width=\linewidth]{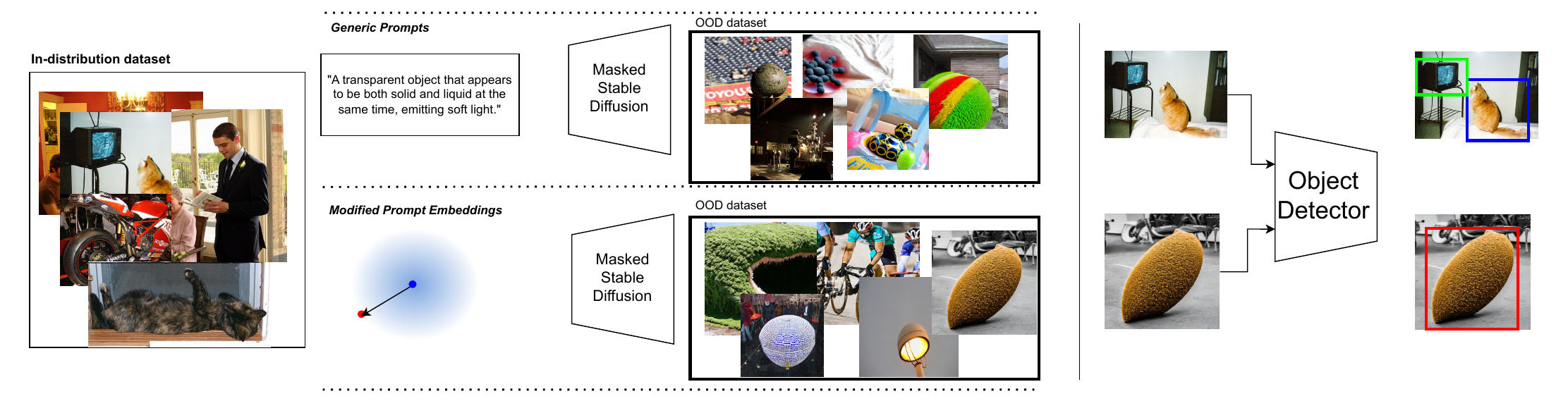}
\caption{Dream-Box overview. We generate outlier objects using two prompt strategies and leveraging in-distribution objects. Subsequently, we train an object detector with a classification output head for in-distribution/OOD.}
\label{fig:overview}
\end{figure*}

On the other hand, Dream-OOD \cite{du2023dream} uses diffusion models to generate full-image synthetic outliers in the pixel space, offering improved OOD detection performance and interpretable visualizations, enhancing the robustness of OOD detectors. Nonetheless, its application to object detection via object-wise outlier generation remains an area of investigation and hence forms the focus of the  OOD study presented here in this paper.

\section{Dream-Box}
In general, object-wise OOD refers to detecting objects in an image and labeling them as either an in-distribution class or an OOD instance, in a semi-supevised mannner, without having any ground-truth (true) OOD samples available during training. Therefore, we introduce Dream-Box, a framework for object bounding-box based OOD that leverages diffusion models for object-wise outlier generation in the pixel space. We develop our strategy for outlier synthesis in \cref{sec:outlier-generation}, while we outline our OOD detection technique in \cref{sec:ood-det}. Finally, the implementation details are discussed in \cref{sec:implementation}.

\subsection{Outlier Generation Strategy} \label{sec:outlier-generation}

Motivated by Dream-OOD \cite{du2023dream}, Dream-Box uses Stable Diffusion \cite{rombach2022high} to generate object-wise outliers in the pixel space, enabling object-aware OOD detection. An overview of Dream-Box is presented in \cref{fig:overview}. 

Our method consists in augmenting the training dataset by replacing in-distibution objects with synthetic outliers that are labeled as OOD. Specifically, we consider an image $\mathbf{x}$ drawn from the in-distribution $\mathbf{x} \sim \mathcal{D}_\mathit{in}$ and consider its ground truth bounding boxes $\mathcal{B}=\left\{b_k\right\}_{k=1}^M$, where $M$ is the number of ground truth bounding boxes in the image, and each bounding box consists of a class label $c_k$ and four position values, \ie, the top-left coordinates $(x_k, y_k)$ and the width and height $(w_k, h_k)$. An image $\tilde{\mathbf{x}}$ with OOD objects is generated by inpainting an object for each $b_k \in \mathcal{B}$ using a masked generator model $f$ that generates an object given a mask $m_k$ and a prompt $\rho(c_k)$ in function of the class label, such that $\tilde{\mathbf{x}}(b_k) = f(\mathbf{x}, \rho(c_k), m_k)$. In this sense, we take the mask as the region described by the bounding box. If we consider all objects within the input image, then the generated OOD image consists of generating one object at a time in a sequential manner: 
\begin{equation}
    \label{eq:gen-process}
    \tilde{\mathbf{x}}_i = f(\tilde{\mathbf{x}}_{i-1}, \rho(c_i), m_i)\,.
\end{equation}
Dream-Box samples $N$ images from $\mathcal{D}_\mathit{in}$ and uses the process from \cref{eq:gen-process} to generate an OOD dataset $\mathcal{D}_\mathit{ood}$ which will enable object-wise OOD detection (\cref{sec:ood-det}). 
\\
\\
\noindent
We consider two prompting strategies to use with \cref{eq:gen-process}, namely \textit{generic textual prompts} and \textit{distance-based modified prompts}, described as follows:

\noindent
\textbf{Generic textual prompts}: Since the goal is to perform OOD detection while still having a good performance in the in-distribution task (object detection), we propose a simple sampling mechanism that create objects similar to the in-distribution dataset but with features that make them OOD. Therefore, we use generic prompts taking the class name and describing unrealistic and/or impossible characteristics. We choose 20 different prompts, given in \cref{table:prompts}, where \{\} indicates the class name. The motivation of this strategy is that we can learn to recognize objects that do not look like the \textit{normal} as in the in-distribution dataset but can be still detected as objects since they might look like the in-distribution data.

\begin{table*}[t!]
\centering
\begin{tabular}{c|l}
\hline
\textbf{No.} & \textbf{Prompt} \\ \hline
1 & A \{\} that defies the laws of physics, floating in mid-air with strange edges. \\ \hline
2 & A mechanical \{\} with organic, plant-like growths intertwining through its structure. \\ \hline
3 & A transparent \{\} that appears to be both solid and liquid at the same time, emitting soft light. \\ \hline
4 & A \{\} that changes its shape continuously, with shifting low contrast colors and textures. \\ \hline
5 & A futuristic \{\} that blends digital and physical elements, glowing with an otherworldly light. \\ \hline
6 & A \{\} with an impossible texture, smooth like liquid but solid like metal, floating in space. \\ \hline
7 & A \{\} that merges two unrelated materials, seamlessly integrating them in an abstract form. \\ \hline
8 & A floating \{\} with intricate geometric patterns constantly changing on its surface. \\ \hline
9 & A \{\} made of plastic that constantly reconfigures itself into different shapes. \\ \hline
10 & A \{\} that appears to be in multiple states at once, existing in two places simultaneously. \\ \hline
11 & A strange \{\} that casts light in ugly but usual colors, transforming its appearance as it moves. \\ \hline
12 & A \{\} that is both solid and ethereal, with strange veins of energy running through it. \\ \hline
13 & A \{\} suspended in time, frozen in mid-motion, with particles of light trailing behind it. \\ \hline
14 & A mysterious floating \{\} with a strange core, surrounded by shifting shadows. \\ \hline
15 & A \{\} made of multiple contrasting materials that somehow coexist harmoniously. \\ \hline
16 & A complex \{\} with multiple layers, each one having a different texture and ugly color that shifts over time. \\ \hline
17 & A \{\} that seems to have multiple dimensions, existing in more than one space at once. \\ \hline
18 & A \{\} with a constantly rotating surface, covered in strange markings and symbols. \\ \hline
19 & A \{\} that looks like it's part of the natural world, but is made entirely of artificial materials. \\ \hline
20 & A smooth \{\} that seems to be melting and reforming simultaneously, surrounded by mist. \\ \hline
\end{tabular}
\caption{List of Prompts for the Generic Prompts strategy. \{\} indicates the name of a class.}
\label{table:prompts}
\end{table*}

\noindent
\textbf{Distance-based modified prompts}: Following Dream-OOD, this strategy aims to perturb the text embedding of the class name in a region relatively far from the embedding. Therefore, given a class name $c_k$, we first obtain its text embedding $\zeta(c)=\mathrm{CLIP}(c)$ and perturb it using random noise, such that the prompt embedding becomes:
\begin{equation}
    \label{eq:std}
    \rho(c)=\zeta(c) + \sigma \epsilon\,,
\end{equation}
where $\epsilon \sim \mathcal{N}(\boldsymbol{0}, I)$ and $\sigma$ is the standard deviation. The aim of this strategy is to create near-anomalies that, while similar to the in-distribution, serve as cues as what means to be a normal object instance. Compared with the generic textual prompts strategy, this method does not require an explicit description of what an OOD object should look like, then avoiding to introduce any bias in the object detector. 

\subsection{Out-of-distribution Detection} \label{sec:ood-det}
\begin{figure}[!t]
\centering
\includegraphics[width=\linewidth]{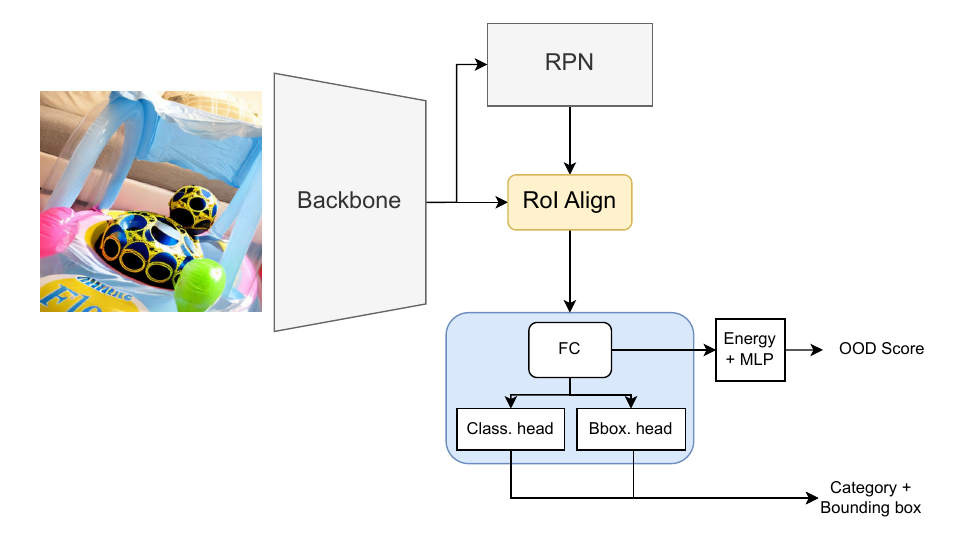}
\caption{Object detector with modified OOD output (head).}
\label{fig:detector}
\end{figure}

The Dream-Box framework enables an augmented dataset $\mathcal{D} = \mathcal{D}_\mathit{in} \cup \mathcal{D}_\mathit{ood}$ with images containing in-distribution and OOD objects. Therefore, any standard object detector can be used for the in-distribution object recognition. With regards to the OOD object detection task, a naive approach of simply learning OOD objects as belonging to an additional class, we aim to detect unseen objects as predict them as OOD. Therefore, a supervised approach as such might not generalize to unseen instances since they might look quite different form the generated OOD instances. In this sense, energy-based methods have shown to have a good performance for OOD in object detection \cite{du2022vos,kumar2023normalizing,isaac2024towards}. Specifically, the energy score of an object feature representation $\mathbf{v}$ is given by:
\begin{equation}
    E(\mathbf{v}) = -\log \sum_{i=1}^K \exp (g_k(\mathbf{v}))\,,
\end{equation}
where $K$ is the number of classes and $g_k(\mathbf{x})$ is the logit of the $k$-th class. Then, this energy is passed to a small multi-layer perceptron $\phi(E)$ that is trained for in-distribution/OOD classification using binary cross entropy, such that we add the extra loss term to the object detector:
\begin{equation}
    \begin{split}
        \label{eq:ood-loss}
        \mathcal{L}_\mathit{ood} = \mathbb{E}_{\mathbf{v}_\mathit{in} \sim \mathcal{D}_\mathit{in}} \left[-\log \frac{e^{\phi(E(\mathbf{v}_\mathit{in}))}}{1 + e^{\phi(E(\mathbf{v}_\mathit{in}))}}\right] + \\ \mathbb{E}_{\mathbf{v}_\mathit{ood} \sim \mathcal{D}_\mathit{ood}} \left[-\log \frac{1}{1 + e^{\phi(E(\mathbf{v}_\mathit{ood}))}}\right]\,.
    \end{split}
\end{equation}
The outline of this approach is shown in \cref{fig:detector}.

\begin{table*}[ht!]
\centering
\begin{tabular}{l|c|c|c}
\hline
\textbf{Method} & \textbf{FPR95 (\%)} & \textbf{AUROC (\%)} & \textbf{mAP (ID) (\%)} \\ \hline
MSP \cite{hendrycks2016baseline} & 70.99 & 83.45 & 48.7 \\
ODIN \cite{liang2017enhancing} & 59.82 & 82.20 & 48.7 \\
Mahalanobis \cite{mahalanobis} & 96.46 & 59.25 & 48.7 \\
Energy score \cite{liu2020energy} & 56.89 & 83.69 & 48.7 \\
Gram matrices \cite{sastry2020detecting} & 62.75 & 79.88 & 48.7 \\ 
Generalized ODIN \cite{hsu2020generalized} & 59.57 & 83.12 & 48.1 \\ 
VOS \cite{du2022vos} & 47.77 & 89.00& 51.5 \\ 
FFS \cite{kumar2023normalizing} & 44.15 & 89.71 & 51.8 \\ \hline
\textbf{Generic Prompts} (\textit{Ours}) & 59.37 & 80.43 & 48.2 \\ 
\textbf{Distance-based modified embeddings} (\textit{Ours}) & 65.03 & 79.27 & 49.5 \\ \hline
\end{tabular}
\caption{Performance comparison of Dream-Box to current OOD techniques for object detection.}
\label{tab:results}
\end{table*}

For this work, we use Faster R-CNN \cite{ren2016faster} with a ResNet50 \cite{he2016deep} backbone. Following previous works in energy-based OOD detection \cite{du2022vos,kumar2023normalizing,liu2020energy,du2023dream,isaac2024towards}, the feature representation of an object $\mathbf{v}$ comes from the penultimate layer of the object classification head. Additionally, since the generated outliers are still intended to be detected as objects, the Region Proposal (sub-)Network within the Faster RCNN architectire is trained with these instances as objects (therefore, increasing their objectness score). Nevertheless, to avoid a negative impact on the in-distribution task, the classification and bounding box heads are not trained for OOD instances.

\subsection{Evaluation and Implementation details}  \label{sec:implementation}
\textbf{Diffusion model}. We use Stable Diffusion fine-tuned for image inpainting. Specifically, the Stable Diffusion v2 model \cite{rombach2022high} is fine-tuned for 200k epochs using the LaMa strategy for masked generation \cite{suvorov2022resolution}. Regarding the \textit{distance-based modified prompt} strategy, choosing a standard deviation value in \cref{eq:std} would require us to know the how far the OOD objects are, which is not known \textit{a priori}. Therefore, we try different standard deviation values, such that  $\sigma \in \left\{0.01, 0.1, 1.0, 2.5, 5.0\right\}$. We sample $N = 5,000$ images with repetition for each experiment. Finally, since we observed that Stable Diffusion v2 might erase objects when the mask area is too small, we only synthesize new objects whose bounding boxes have an area $A >2,000$ pixels.

\noindent
\textbf{Object Detection}. We use MMDetection \cite{chen2019mmdetection} for training Faster RCNN \cite{ren2016faster} with a ResNet50 \cite{he2016deep} backbone pretrained on the ImageNet \cite{imagenet}. All of our models are trained using Stochastic Gradient Descent for 18 epochs (following Du \etal \cite{du2022vos}), with a batch size of 16, weight decay of $1\times 10^{-5}$ and initial learning rate of 0.02 that is decreased by a factor of 0.1 after 12 and 16 epochs. The OOD classifier is trained using Focal Loss \cite{lin2017focal} with a loss weight of 10.0. Each model is trained using a single NVIDIA A100 GPU.

\noindent
\textbf{Datasets}. We use the PASCAL VOC 2007/12 dataset \cite{voc}, comprising 20 object classes, as the in-distribution dataset. The OOD dataset consists of the testing partition of the MS-COCO dataset \cite{lin2014microsoft}, removing all images that contain any of the 20 in-distribution classes, with the final OOD testing dataset consisting of 930 images. 

\noindent
\textbf{Evaluation Metrics}
We report the area under the receiver operating characteristic (AUROC) curve for OOD detectetion, and the false positive rate at 95\% confidence (FPR95) of in-distribution detection, \ie, the rate of OOD instances labeled as in-distribution given an OOD score threshold such that 95\% of in-distribution instances are correctly classified. Additionally, we report average precision (AP) in the in-distribution object detection task to evaluate the effect of Dream-Box in the main task. 

\section{Results}

\begin{figure}[!t]
\centering
\includegraphics[width=\linewidth]{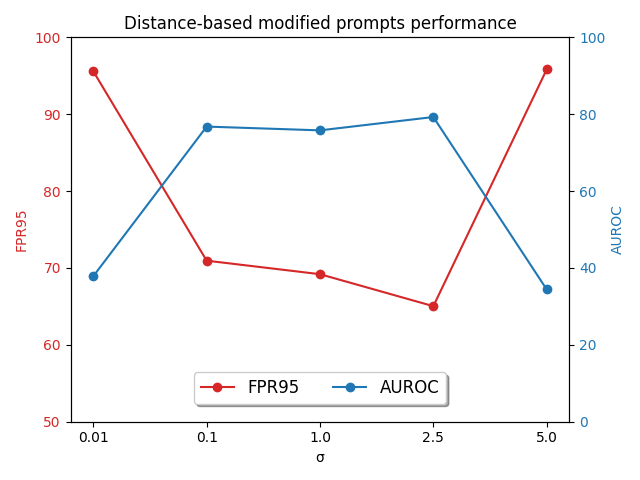}
\caption{Performance of the distance-based modified prompting strategy with different $\sigma$ values.}
\label{fig:ablations}
\end{figure}

\begin{figure*}[!t]
\centering
\includegraphics[width=0.95\linewidth]{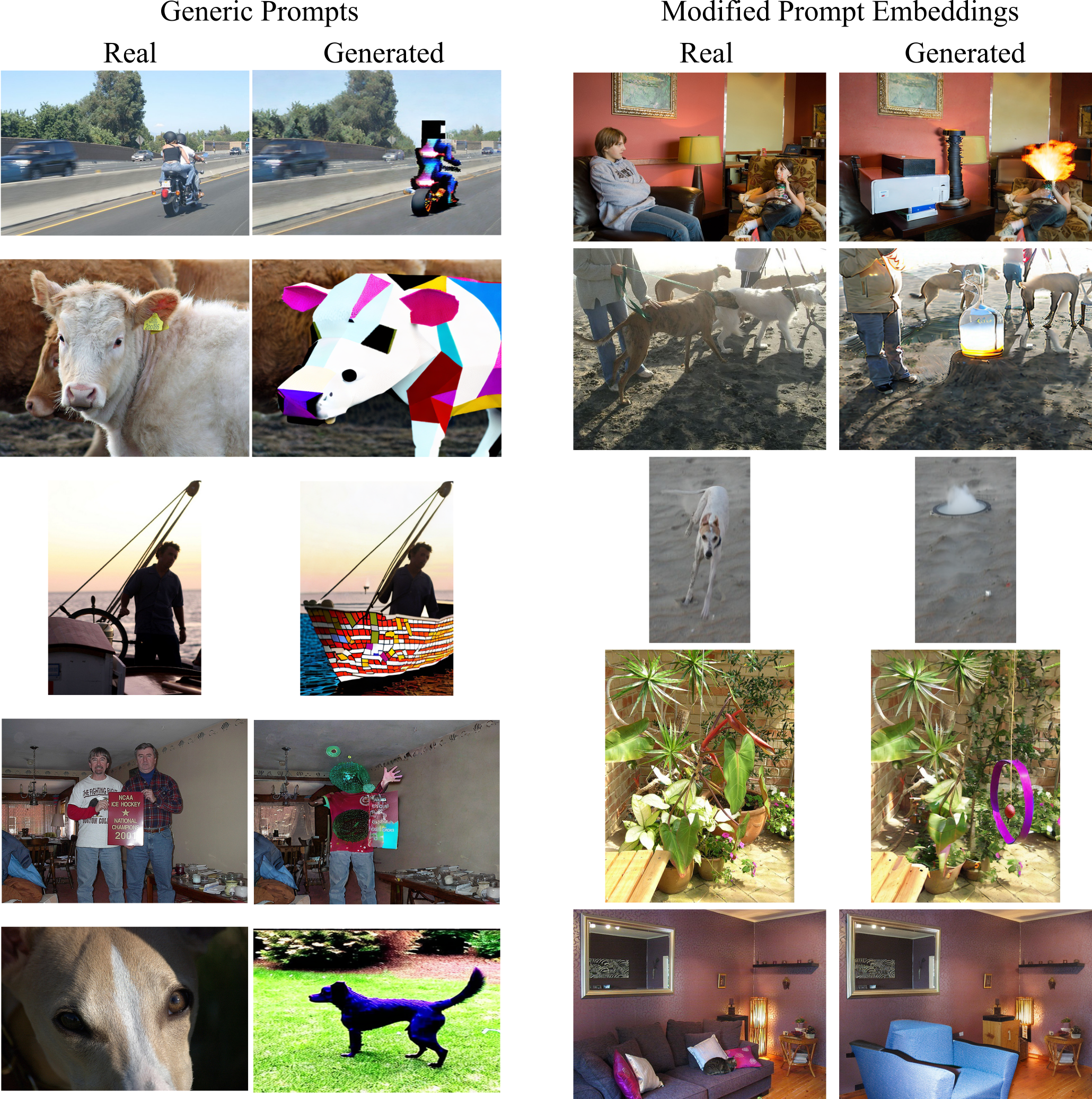}
\caption{Exemplar generated outlier objects of the generic prompt and distance-based modified prompting strategies.}
\label{fig:outliers}
\end{figure*}

\noindent
\cref{tab:results} shows the results of our method compared to other standard OOD approaches in object detection. The results shown for the distance-based modified prompt embeddings, corresponding to $\sigma=2.5$, achieved the best performance. Although our method has a lower performance compared with the state-of-the-art methods (VOS and FFS), it still achieves comparable performance with respect to other traditional approaches, indicating that the generated outliers are providing cues into what are in-distribution samples or not. A key principle in VOS and FFS is that the outliers are generated from the feature representation, therefore learning a compact representation of the feature space and classifying other instances as OOD. This indicates that the feature representation of the objects in Faster RCNN has strong signals of whether an object is part of the distribution or not. However, our method provides a sense of explainability by enabling visualization of outliers in the pixel space. \cref{fig:outliers} shows the generated outliers for both strategies. The improved performance of the \textit{generic prompts} is directly related to the outliers being closer to the actual object distribution than the \textit{distance-based modified prompt embeddings} approach. For instance, the dog in the third row completely disappears, indicating that the $\sigma$ value used to generate the embeddings is far from the class embedding. Therefore, the generated images from the \textit{generic prompts} approach indicate that showing images near the in-distribution data but with OOD features helps in OOD detection for object detection.

\begin{figure*}[!t]
\centering
\includegraphics[width=\linewidth]{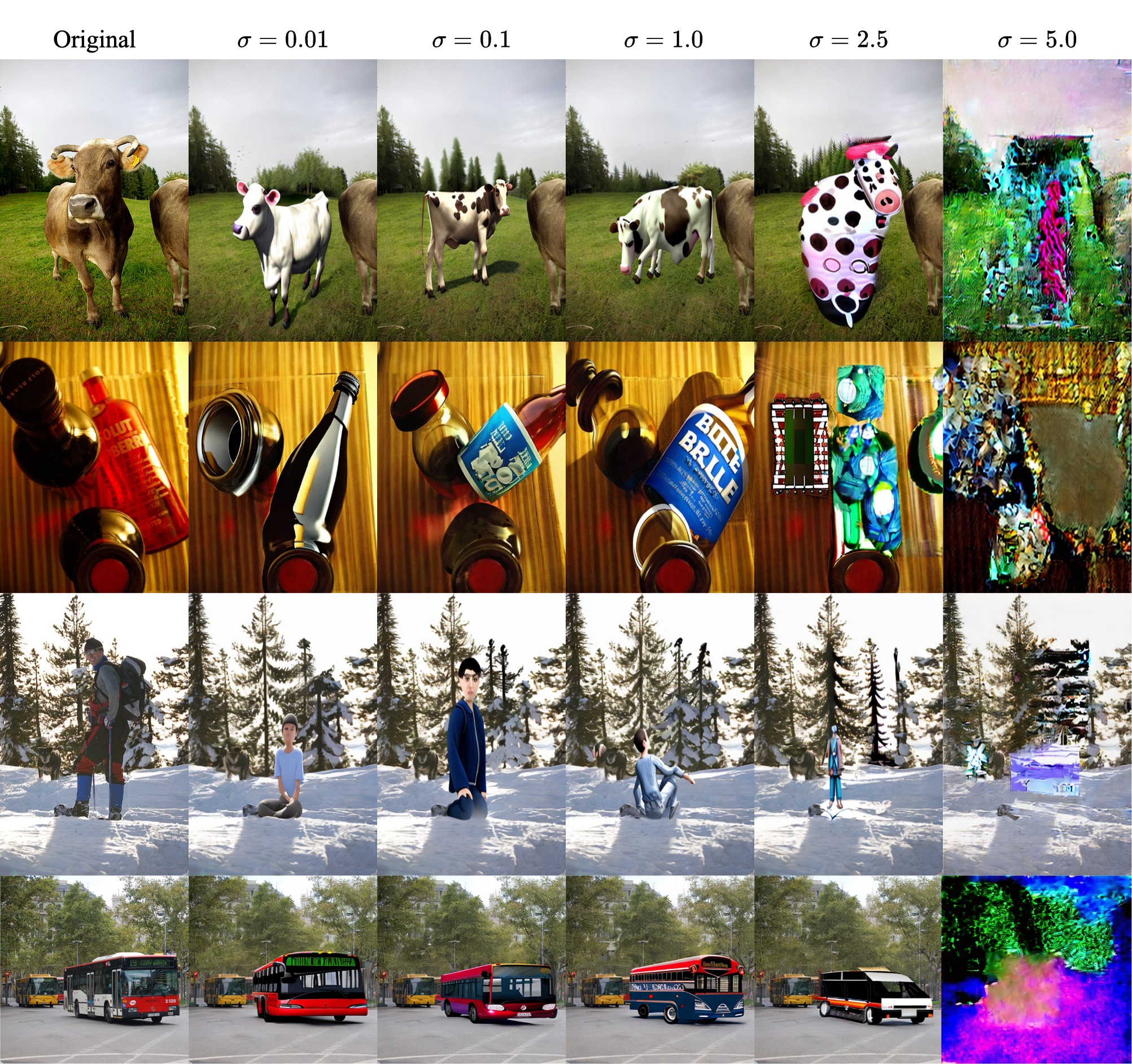}
\caption{Outlier objecr generation comparison by varying $\sigma$ in the distance-based modified prompting strategy.}
\label{fig:sigma-outliers}
\end{figure*}

Ablation studies corresponding to different values of $\sigma$ in the distance-based modified prompt strategy are presented at \cref{fig:ablations}, where it is observed that the best performance is achieved at $\sigma=2.5$. The ablations also show that using too small or too large values is detrimental to Dream-Box performance. For small $\sigma$ values, the generated outliers might look quite similar to the original in-distribution data, which is opposite to the goal of generating synthetic outliers. On the other hand, too large values of $\sigma$ create very dissimilar objects that might be too far from the object distribution (\ie, they do not look like objects anymore). 

This is further supported by \cref{fig:sigma-outliers}, where we show different outliers by varying $\sigma$ values. For instance, most of the objects with $\sigma \leq 1.0$ are similar to what in-distribution objects look like, whereas for $\sigma=5.0$ the images break and no longer show identifiable objects. On the other hand, while objects generated with $\sigma=2.5$ can still be identified as an object with some resemblance to the original class, they present anomalous features that make them ideal candidates for OOD training. Nonetheless, it is also observed that objects might look still like the original class in some instances (such as the bus in the last row), or completely unrecognizable (\eg, the bottle in the second row). This indicates that identifying the proper distance to the class-embedding centre for outlier synthesis might be critical for proper class-aware OOD object synthesis. 

Recent works \cite{du2023dream,tao2023nonparametric} have shown other strategies for OOD sampling without relying to explicit distances from the class name embedding, but their application to object-wise ODD are as yet un-investigated. Therefore, the study of improved outlier sampling strategies for object-wise OOD remains an area for future work.
\section{Conclusion}

This work introduces Dream-Box, a novel framework for outlier object generation in the pixel space for object-wise out-of-distribution (OOD) detection. By leveraging diffusion models, our method synthesizes pixel-space outliers, addressing the lack of interpretability which is the key limitation of alternative feature-space approaches. Unlike prior methods such as VOS \cite{du2022vos} and FFS \cite{kumar2023normalizing}, which rely on abstract feature representations, Dream-Box provides a more intuitive and explainable means of generating synthetic outliers, thereby enhancing the explainability of OOD detection models.

Experimental evaluations demonstrate that Dream-Box achieves competitive OOD detection performance with traditional OOD approaches, despite not surpassing state-of-the-art feature-space methods. Notably, the generic prompt strategy yields improved OOD classification results compared to distance-based modified prompt embeddings, suggesting that generating outliers closer to the decision boundary contributes positively to detection accuracy. The ability to visualize synthetic outliers offers additional insights into the failure modes of object detection models, an aspect previously unexplored in pixel-space OOD generation for this task. Further analysis shows that while the distance-based modified prompt strategy for outlier generation underperforms the generic prompt strategy, it provides a tunable parameter for controlling the objects anomalous appearance. Additionally, the use of improved sampling methods in the class embedding space may improve such a strategy, although its application to object detection remains unexplored.

These findings underscore the potential of pixel-space outlier generation for interpretable OOD detection in object detection. Future research directions include refining the generation process to better balance in-distribution performance with OOD separability, extending Dream-Box to additional vision tasks, and exploring more adaptive prompt engineering techniques to improve the quality of synthesized outliers, including the exploration of the text embedding space for OOD prompt embeddings, akin Dream-OOD \cite{du2023dream}.
{
    \bibliographystyle{ieeenat_fullname}
    \bibliography{main}
}


\end{document}